\documentclass{article}

\usepackage{amsmath} 
\usepackage{amsthm}
\usepackage{comment}

\usepackage{tcolorbox}

\usepackage[preprint]{neurips_2026}

\usepackage[utf8]{inputenc} %
\usepackage[T1]{fontenc}    %
\usepackage{hyperref}       %
\usepackage{url}            %
\usepackage{booktabs}       %
\usepackage{amsfonts}       %
\usepackage{nicefrac}       %
\usepackage{microtype}      %
\usepackage{xcolor}         %
\usepackage{graphicx}
\usepackage{xspace}
\usepackage[english]{babel}
\usepackage{multirow}
\usepackage{bbding}

\newcommand{\ie}{i.e.,\@\xspace}
\newcommand{\eg}{e.g.,\@\xspace}
\newcommand{\etc}{etc.\@\xspace}

\newcommand{\greencheck}{\textcolor{green}{\Checkmark}}
\newcommand{\cross}{\textcolor{red}{\XSolid}}
\newcommand{\Hedge}{$\mathcal{H}_{\text{edge}}$}
\newcommand{\Hguidecode}{$\mathcal{H}_{\text{code}}^{\text{guided}}$}
\newcommand{\Hopencode}{$\mathcal{H}_{\text{code}}^{\text{open}}$}

\newcommand{\methodname}{DiscoPER\xspace}

\title{Autonomous Scientific Discovery \\ via Iterative Meta-Reflection}

\author{%
  Bingchen Zhao\textsuperscript{1} \quad
  Sara Beery\textsuperscript{2} \quad
  Oisin Mac Aodha\textsuperscript{1}\\
  \textsuperscript{1}University of Edinburgh \quad \quad
  \textsuperscript{2}Massachusetts Institute of Technology
}

\begin{document}

\maketitle

\begin{abstract}
Autonomous scientific discovery systems offer the potential to accelerate research by automating the process of hypothesis generation and validation. 
However, current systems operate within constrained search spaces or require predefined research questions, limiting their capacity for true open-ended inquiry. 
Furthermore, while they generate hypotheses iteratively, they largely lack the ability to explicitly synthesize their own accumulated findings to uncover complex, interconnected phenomena. 
We introduce \methodname, an autonomous large language model-powered framework that conducts open-ended research by dynamically generating and executing code to explore datasets without pre-specified research objectives. 
To ensure rigorous scientific validity, every proposed discovery must pass statistical testing.
To overcome the limitations of isolated search, our framework introduces a second-order reasoning mechanism that periodically analyzes its own accumulated discoveries. 
By treating prior discoveries as empirical data, \methodname identifies structural patterns, confounds, and epistemic gaps, actively redirecting hypothesis exploration toward uncharted regions of the search space. 
The search space is further expanded by incorporating tool use, enabling the system to explore hypotheses beyond structured metadata by seamlessly processing and extracting useful information from multimodal sources like images.
Evaluated on iNatDisco, a new multimodal ecological  knowledge benchmark   with pattern-level ground truth obtained from peer-reviewed literature, \methodname recovers 8 of 9 known patterns with a 72.7\% hypothesis support rate, outperforming both classical causal discovery and  LLM-guided baselines.
Ablations show that \methodname scales with more data, and confirms the benefits of  second-order ``meta-reflection''. 
\end{abstract}

\section{Introduction}

Scientific discovery is a cumulative process. 
Researchers build on prior findings, notice gaps in what has been tested, and revise theories based on accrued evidence. 
Most existing LLM-based discovery systems (\eg \citep{panigrahi2026heurekabench,gupta2026accelerating,gottweis2025ai_coscientist}) treat each hypothesis generation step independently, \ie the system is provided with a dataset or research question, it generates a hypothesis, and the loop resets. 
There is no mechanism for the system to survey what it has already learned and reason about what remains unknown. 
A human biologist, for instance, might notice that several pairwise correlations all involve the same variable and hypothesize a common cause, or review hundreds of field photographs to identify a seasonal migration pattern that no single observation reveals.
This meta-level reflection across findings is central to science, yet is an aspect current discovery systems have largely ignored. %

Moreover, existing systems are either restricted in what they can express or require external guidance to operate.
Classical structure learning methods~\citep{spirtes2000causation_PC,zheng2018NOTEARS,chickering2002optimal_GES} search exhaustively over pairwise variable relationships but cannot express higher-order patterns such as interactions, mediation chains, or confound structures.
Full-pipeline AI scientist systems~\citep{lu2024ai_scientist,gottweis2025ai_coscientist,ghareeb2025robin} can generate rich hypotheses from literature and domain knowledge, but require initial research objectives or seed hypotheses to guide their search.
Code-based concurrent works~\citep{gupta2026accelerating,panigrahi2026heurekabench} write executable analyses, but their search is constrained by externally-supplied questions or task summaries.
None of these systems perform fully open-ended discovery, \ie (i) starting from raw unstructured data alone without a pre-specified question, (ii) expressing arbitrary testable hypotheses as executable code, and (iii) autonomously deciding what to investigate next based on what has already been discovered.

\begin{figure}[t]
\centering
\includegraphics[width=1.0\textwidth]{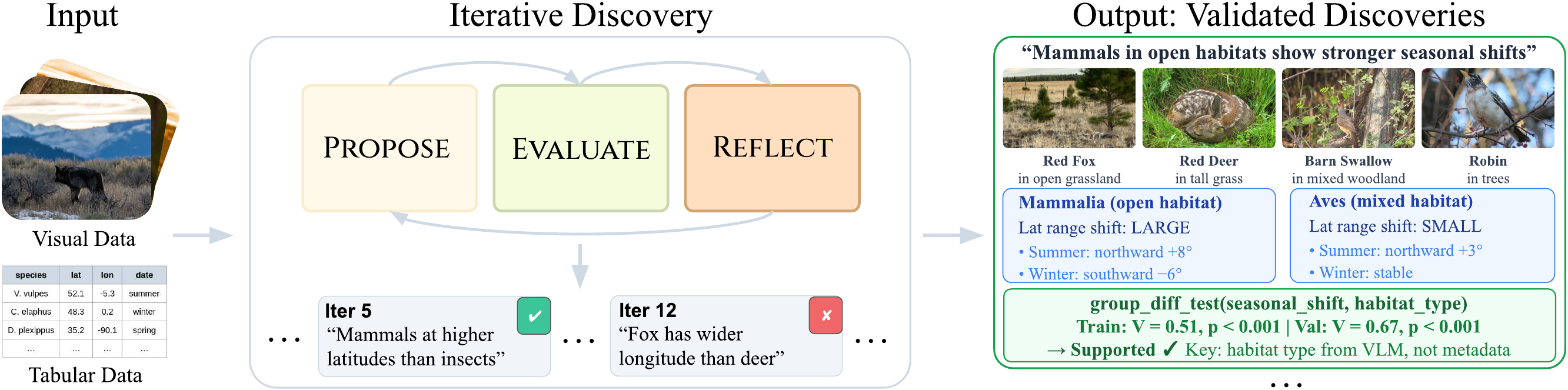}
\vspace{-1.5em}
\caption{We introduce \methodname{}, an iterative approach for autonomous scientific discovery that takes multimodal data as input and generates a set of validated  discoveries pertaining to the input data as output. 
At each iteration, the system proposes hypotheses, executes statistical tests on the underlying data, and accepts only discoveries that pass held-out validation. 
Periodically, the \textsc{Reflect} module analyzes the accumulated accepted and rejected claims to identify gaps, confounds, and promising compound hypotheses, which then guide the next round of exploration.
}
\label{fig:overview}
\vspace{-1.5em}
\end{figure}

We present \methodname, an autonomous discovery system that satisfies all three requirements (see Fig.~\ref{fig:overview}).
At its core, \methodname contains LLM agents that receive data and propose hypotheses without being told what to look for a priori.
Each hypothesis is realized as executable code that invokes statistical tools %
to ground every claim in a formal test on real data, rather than in LLM confidence estimates~\cite{kiciman2024causal_llm_aug_cd,jiralerspong2024efficient_bfs}.
Claims that pass statistical effect size and significance thresholds %
enter a persistent claim store.
Periodically, a reflection step analyzes the current claims to identify under-explored variables and to detect recurring confounds across them. 
From this, it generates targeted guidance for the next model iteration, all without access to ground truth data.

In summary, we formalize open-ended discovery as a generalized \textsc{Propose}--\textsc{Evaluate}--\textsc{Reflect} loop and show that existing systems are restricted instances of this framework, lacking either flexible hypotheses space, unrestricted hypothesis generation, or reflective iterative search over the hypotheses space.
We instantiate this framework as a multimodal, code-based system in which agents write and execute statistical analyses grounded in real data, validated on held-out splits.
Because existing benchmarks are often restricted by the research questions it asks, we construct new benchmarks from real scientific data where ``ground-truth'' patterns have been identified and the data itself allows open-ended discovery.
We introduce a new multimodal dataset called iNatDisco sourced from public citizen science data~\cite{iNaturalist} that contains expert verified claims that are grounded in the scientific literature. 
In our experiments, we demonstrate that \methodname obtains the best discovery performance across multiple benchmarks and that reflection enhances hypothesis generation by allowing \methodname to better explore the hypotheses space. 
Most importantly, we show that \methodname is able to rediscover over 60\% real world human expert verified patterns on iNatDisco.

\section{Related work} 

\noindent \textbf{Causal discovery and structure learning.}
Our discovery framework bridges classical structure learning and recent LLM-augmented discovery. 
Traditional algorithms, including constraint-based~\citep{spirtes2000causation_PC}, score-based \citep{chickering2002optimal_GES}, and continuous optimization methods \citep{zheng2018NOTEARS}, are fundamentally restricted to searching a predefined space that only contains the edges between variables in a dataset. 
More recently, LLM-based causal discovery methods \citep{kiciman2024causal_llm_aug_cd,jiralerspong2024efficient_bfs,jin2024corr2cause} have emerged to improve search efficiency or inject domain knowledge. 
However, these approaches primarily replace a statistical testing oracle with LLM reasoning prompts or heuristic semantic judgments while remaining rigidly confined to the same edge space. 
Our approach fundamentally departs from these methods by using the LLM strictly as a semantic generator and meta-reasoner to explore the vastly larger and more expressive space of executable hypotheses, while grounding actual verification in formal, data-driven, statistical testing to prevent hallucination.

\noindent \textbf{Autonomous scientific discovery systems.}
Recent systems aspire to automate the full scientific process but differ fundamentally in how they source hypotheses and whether they perform data-driven validation.
The AI Scientist~\citep{lu2024ai_scientist} generates machine learning research ideas, writes code, runs experiments, and produces full papers, but operates on machine learning tasks rather than empirical data analysis and evaluates novelty via simulated review, rather than statistical validation.
The AI Co-Scientist~\citep{gottweis2025ai_coscientist} uses multi-agent debate with tournament evolution to generate biomedical hypotheses that have achieved wet-lab validation, but its hypotheses originate from the literature and domain knowledge, rather than from data-driven statistical analysis.
Robin~\citep{ghareeb2025robin} closes the loop with physical experiments, discovering a novel therapeutic, but is similarly guided by research objectives and literature context.
SciAgents~\citep{ghafarollahi2024sciagents} traverses ontological knowledge graphs to uncover interdisciplinary connections in materials science, generating and critiquing hypotheses through multi-agent reasoning, but does not ground its claims in statistical tests on raw data.
In laboratory settings, the Robot Scientist~\citep{king2004functional_robot_scientist} demonstrated closed-loop hypothesis generation and physical experimentation in yeast genomics, and Coscientist~\citep{boiko2023autonomous_chem} automates chemical synthesis planning and execution with LLM-driven tool use. 
However, both operate in domains with physical feedback loops rather than observational datasets.
BioDiscoveryAgent~\citep{roohani2025biodiscoveryagent} designs gene perturbation panels to achieve target cell phenotypes, but addresses a constrained optimization task rather than open-ended pattern discovery. 

A parallel line of work focuses specifically on LLM-driven hypothesis generation and testing.
\cite{zhou2024hypothesis} propose a framework for generating hypotheses with LLMs, while \cite{agarwal2025autodiscovery} drive discovery through Bayesian surprise, measuring belief shifts after evidence collection.
\cite{wang2024hypothesis} task LLMs with discovering transformation rules from input-output pairs via code execution, and \cite{huang2025automated} identify sub-hypotheses and design falsification experiments executed through code, though their approach remains largely linear and non-iterative. 
Concurrent to our work, ExperiGen~\citep{gupta2026accelerating} unifies hypothesis generation with experimental validation, incorporating statistical tests and a short-term memory module.
While ExperiGen represents a step toward autonomous discovery, it requires descriptive seed hypotheses and task summaries to guide search, and its memory is conditioned implicitly on prior pairs rather than performing structured analysis of what they collectively imply. 

All of the above systems either require external research direction, lack a mechanism for structured reflection over accumulated findings, or both.
Our approach differs in that it operates data-first, grounds every claim in held-out statistical validation, and uses reflective accumulation to reason about what the collection of findings implies for future exploration.

\noindent \textbf{Benchmarks for scientific discovery.}
The shift from static LLM predictions to dynamic, code-driven agents has necessitated new environments and benchmarks for evaluating scientific discovery \citep{luo2025benchmarking}. 
Recent datasets like BioDSA~\citep{wang2025biodsa}, HeurekaBench~\citep{panigrahi2026heurekabench}, BLADE~\citep{gu2024blade}, and DiscoveryBench~\citep{majumder2024discoverybench} provide structured environments to evaluate agentic research capabilities. 
However, these benchmarks typically evaluate whether an agent can answer a predefined research question or navigate a highly constrained scenario.
\methodname is designed for unconstrained, autonomous discovery, and, as a result, standard task-oriented evaluation falls short. 
To address this, we introduce our own open-ended evaluation paradigm utilizing complex multimodal ecological data from the citizen science platform iNaturalist~\citep{iNaturalist} to assess an agent's ability to autonomously formulate, test, and synthesize supported scientific claims from scratch.

\section{Method}
\label{sec:method}

\begin{figure}
\centering
\includegraphics[width=1.0\textwidth]{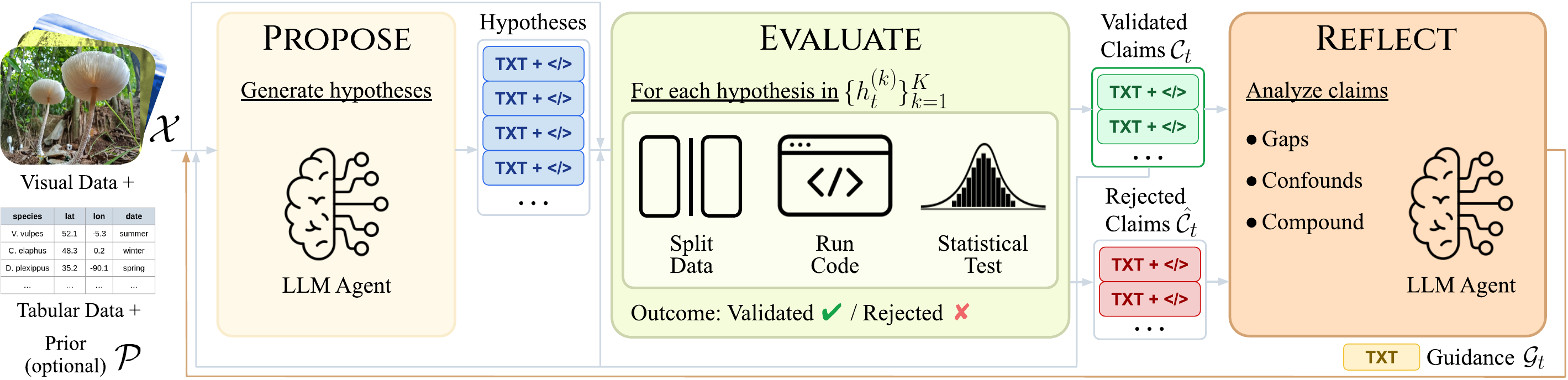}
\vspace{-1.5em}
\caption{\methodname{} is an iterative scientific discovery system consisting of three core modules:  
\textsc{Propose} generates hypotheses based on the data $\mathcal{X}$, and optional prior knowledge $\mathcal{P}$, and generates a set of candidate hypotheses $\{h_t^{(k)}\}_{k=1}^K$, where each is a natural language expression and accompanying code. 
\textsc{Evaluate} generates code to test each hypothesis to either validate or reject them based on statistical evidence supported by the data. 
\textsc{Reflect} analyzes the validated  and rejected claims ($\mathcal{C}_t$ and $\hat{\mathcal{C}}_t$) to produce guidance $\mathcal{G}_t$ which steers the next round of hypothesis generation. 
}
\label{fig:our_apprach}
\vspace{-1.5em}
\end{figure}

We posit that  existing scientific discovery systems can be viewed as restricted instances of a single generalized framework (see Table~\ref{tab:formalization} for a summary).
In this section we introduce each component using our \methodname{} approach, illustrated in Fig.~\ref{fig:our_apprach}, as the primary example.

\begin{table}[t]
\centering
\caption{Existing autonomous scientific discovery systems can be viewed as instances of our generalized framework in Sec.~\ref{sec:method}. $\mathcal{H}$: hypothesis space ($\mathcal{H}_\text{edge}$ = pairwise variable edges, $\mathcal{H}_\text{code}^{\,\text{guided}}$ = executable code with external guidance, $\mathcal{H}_\text{code}^{\,\text{open}}$ = executable code without guidance). \textsc{Propose}: how hypotheses are generated. \textsc{Evaluate}: how hypotheses are tested. \textsc{Reflect}: whether the system reasons over accumulated findings. $\mathcal{P}$: prior information required ($\emptyset$ = none, Partial = task description or seed hypotheses, Full = specific research questions or objectives).}
\label{tab:formalization}
\vspace{-.5em}
\footnotesize
\setlength{\tabcolsep}{4pt}
\begin{tabular}{@{}lccccc@{}}
\toprule
\textbf{Method}
  & $\mathcal{H}$
  & \textsc{Propose}
  & \textsc{Evaluate}
  & \textsc{Reflect}
  & $\mathcal{P}$ \\
\midrule
PC~\citep{spirtes2000causation_PC} / GES~\citep{chickering2002optimal_GES}
  & $\mathcal{H}_\text{edge}$
  & Enum.\ $(V_i,V_j,\mathbf{S})$
  & CI test; $p < \alpha$
  & $\mathcal{G} = \emptyset$
  & $\emptyset$ \\
NOTEARS~\citep{zheng2018NOTEARS}
  & $\mathcal{H}_\text{edge}$
  & Grad.\ desc.\ on $W$
  & LS + DAG pen.
  & $\mathcal{G} = \emptyset$
  & $\emptyset$ \\
GPT-4 BFS~\citep{jiralerspong2024efficient_bfs}
  & $\mathcal{H}_\text{edge}$
  & BFS; LLM judges
  & LLM confidence
  & $\mathcal{G} = \emptyset$
  & Full \\
\midrule
ExperiGen~\citep{gupta2026accelerating}
  & $\mathcal{H}_\text{code}^{\,\text{guided}}$
  & $G_\theta$; refine $\leq\!T$
  & ReAct; Bonferroni
  & Short term memory
  & Partial \\
HeurekaBench~\citep{panigrahi2026heurekabench}
  & $\mathcal{H}_\text{code}^{\,\text{guided}}$
  & $\emptyset$ (given)
  & Agent Eval
  & $\mathcal{G} = \emptyset$
  & Full \\
AI Scientist~\citep{lu2024ai_scientist}
  & $\mathcal{H}_\text{code}^{\,\text{guided}}$
  & LLM ideation
  & ML metrics
  & $\mathcal{G} = \emptyset$
  & Full \\
\midrule
\textbf{\methodname} (Ours)
  & $\mathcal{H}_\text{code}^{\,\text{open}}$
  & $G_\theta(\mathcal{G}_{t\text{-}1},\, \mathcal{C}_{t\text{-}1})$
  & Code
  & $\mathcal{G}_t \!=\! \textsc{Reflect}(\mathcal{C}_t)$
  & $\emptyset$ \\
\bottomrule
\end{tabular}
\vspace{-1.5em}
\end{table}

\noindent \textbf{Setup.}
Let $\mathcal{X}$ be a dataset of $N$ observations, optionally accompanied by additional data such as images.
Let $\mathcal{P}$ denote any information beyond $\mathcal{X}$ available before the discovery loop begins: $\mathcal{P} = \emptyset$ (none), $\mathcal{P} = \text{Partial}$ (e.g., a dataset summary), or $\mathcal{P} = \text{Full}$ (e.g., specific questions to answer).
A \emph{hypothesis} $h$ is a program that takes $\mathcal{X}$ as input and returns a judgment $b \in \{\text{supported}, \text{rejected}\}$ together with statistical evidence $e$.
The \emph{hypothesis space} $\mathcal{H}$ is the set of all such programs the system can express.
A hypothesis that has been executed and accepted is called a \emph{claim} / \emph{discovery}. 
The \emph{claim set} $\mathcal{C}_t \subseteq \mathcal{H}$ collects all claims accepted after $t$ iterations, and we use $\hat{\mathcal{C}_t}$ to denote the set of  rejected claims at $t$.
The \emph{claim set} on its own can accumulate discovered patterns, but it cannot conduct a meta-level analysis and use the accumulated knowledge to guide the next step of discovery.
To address this, additional \emph{guidance} is needed. 
The \emph{guidance} $\mathcal{G}_t$ is a structured summary produced by analyzing $\mathcal{C}_t$ and $\hat{\mathcal{C}_t}$, operating at a higher level than individual claims. 
Specifically, it describes what has been explored, what is missing, and what to try next.

\noindent \textbf{The discovery loop.}
Starting from $\mathcal{C}_0 = \emptyset$ and $\mathcal{G}_0 = \emptyset$, the system iterates:
\begin{align}
h_t^{(1)}, \ldots, h_t^{(K)} &\sim \textsc{Propose}(\mathcal{X},\; \mathcal{C}_{t-1},\; \mathcal{G}_{t-1},\; \mathcal{P}) \label{eq:propose} \\
\Delta \mathcal{C}_{t}, \Delta \hat{\mathcal{C}}_{t} &= \textsc{Evaluate}(\{h_t^{(k)}\}_{k=1}^K,\; \mathcal{X},\; \mathcal{C}_{t-1},\; \hat{\mathcal{C}}_{t-1}) \label{eq:evaluate_delta} \\
\mathcal{C}_t= \mathcal{C}_{t-1} &\cup \Delta \mathcal{C}_{t}, \;
\hat{\mathcal{C}}_{t}
= \hat{\mathcal{C}}_{t-1} \cup \Delta \hat{\mathcal{C}}_{t} \label{eq:update_claims} \\
\mathcal{G}_{t} &= \textsc{Reflect}(\mathcal{C}_{t}, \hat{\mathcal{C}_t}) \label{eq:reflect}
\end{align}

At each iteration, \textsc{Propose} generates $K$ hypotheses conditioned on the current claims $\mathcal{C}_{t-1}$ and guidance $\mathcal{G}_{t-1}$.
\textsc{Evaluate} tests each hypothesis and adds those that pass to the claim set.
\textsc{Reflect} then analyzes the updated claims to produce updated guidance $\mathcal{G}_t$, which steers the next round of hypothesis proposal using \textsc{Propose}.
Systems without reflective reasoning set $\mathcal{G}_t = \emptyset$ for all $t$.

\noindent {\textbf{\textsc{Propose}:} multimodal hypothesis generation.}
An LLM agent receives a tabular summary of $\mathcal{X}$ (schema, column statistics, sample rows) and, when available, additional metadata such as images. %
Conditioned on this input, the current claim set $\mathcal{C}_{t-1}$ and the guidance $\mathcal{G}_{t-1}$, the agent outputs $K$ structured hypotheses, each consisting of a natural-language statement, the variables involved, and accompanying Python code that implements a statistical test to verify the hypothesis.
The guidance $\mathcal{G}_{t-1}$ steers \methodname{} toward under-explored regions of the hypothesis space without dictating specific hypotheses.
By default, \methodname{} operates with  $\mathcal{P} = \emptyset$, \ie  it is not told what to look for. %

\noindent {\textbf{\textsc{Evaluate}:} code execution and held-out validation.}
Each of the $K$ hypothesis programs is executed on a training split of $\mathcal{X}$, invoking statistical tests such as correlation analysis, group comparison, predictive modeling, clustering with enrichment testing, or stratified subgroup re-analysis.
The same code is then re-executed on a held-out validation split.
A hypothesis is accepted into $\mathcal{C}_t$ only if the effect size exceeds a minimum threshold and the $p$-value falls below a significance level on both train and validation splits.
The system is able to tune the code used for validating the hypothesis on a training set, but can only evaluate the code once on the validation set.
This design is essential to avoid p-hacking where the system collect data or refines the analysis until nonsignificant results become significant~\cite{head2015extent_phack}.
As \methodname{} can write arbitrary Python code, its hypothesis space $\mathcal{H}_\text{code}^{\,\text{open}}$ is in principle the space of all Turing-computable statistical tests.
This contrasts with classical causal discovery methods (\eg~\citep{zheng2018NOTEARS,chickering2002optimal_GES,spirtes2000causation_PC}), which are restricted to $\mathcal{H}_\text{edge}$, \ie programs that test a single directed edge $V_i \to V_j$ via a fixed conditional independence test, yielding $|\mathcal{H}_\text{edge}| = O(d^2)$.
Code-based concurrent works such as ExperiGen~\citep{gupta2026accelerating} and HeurekaBench~\citep{panigrahi2026heurekabench} also write programs, but they require an externally-supplied research question or target pattern ($\mathcal{P} \neq \emptyset$): HeurekaBench evaluates agents on pre-specified research questions paired with ground-truth answers, and ExperiGen similarly operates on user-specified hypotheses.
Without such external guidance these systems have no basis for selecting what to investigate, guidance is a structural requirement for these systems.
Our system can also accept optional guidance, but it is designed to operate with $\mathcal{P} = \emptyset$, freely exploring the full open space.
We denote the constrained space $\mathcal{H}_\text{code}^{\,\text{guided}}$.
These spaces form a hierarchy: $\mathcal{H}_\text{edge} \subset \mathcal{H}_\text{code}^{\,\text{guided}} \subseteq \mathcal{H}_\text{code}^{\,\text{open}}$.

\noindent {\textbf{\textsc{Reflect}:} meta-level guidance.}
Without explicit guidance, the hypothesis generator tends to revisit variants of previously explored hypotheses.
Our \textsc{Reflect} module addresses this by periodically analyzing the full claim set to redirect the search.
After each round of evaluation, a separate LLM agent receives the accepted claims $\mathcal{C}_t$ and rejected claims $\hat{\mathcal{C}}_t$ and produces structured guidance $\mathcal{G}_t$.
This guidance is a piece of text that prompts \textsc{Propose} to explore different aspects of the dataset.
In practice, the guidance takes diverse forms depending on the state of the claim store.
It may identify \emph{gaps}, where variables or relationships that are underexplored in $\mathcal{C}_t$.
It may flag \emph{confounds}, where a variable appears as a moderator across multiple accepted claims, suggesting that existing findings may be driven by an uncontrolled factor.
Or it may propose \emph{compound hypotheses}, where two or more accepted claims share overlapping variables in ways that imply a higher-order relationship not yet tested.
At the next iteration, \textsc{Propose} conditions on $\mathcal{G}_t$, as is steered towards these opportunities without dictating specific hypotheses.
\textsc{Reflect} has no access to ground truth and reasons entirely from the system's own output.

This explicit reflection over accumulated evidence distinguishes \methodname{} from concurrent work: ExperiGen~\citep{gupta2026accelerating} conditions on prior hypotheses individually but never analyzes them collectively, and HeurekaBench~\citep{panigrahi2026heurekabench} has no mechanism to revise the search direction based on its findings.
Our \methodname approach, displayed in Fig.~\ref{fig:our_apprach}, is the first to combine $\mathcal{H}_\text{code}^{\,\text{open}}$, $\mathcal{P} = \emptyset$, and use explicit reflection. 
This flexibility makes it well suited for open-ended discovery.

\section{Experiments} 
Here we present quantitative and qualitative results comparing \methodname to alternative autonomous scientific discovery approaches. 
Results on additional datasets can be found in Appendix~\ref{sec:additional_results}. 

\subsection{Implementation details}
\label{sec:implementation}

\noindent \textbf{Models.}  
We evaluate \methodname with Claude Sonnet 4.6 as the default LLM, and additionally report results with Claude Opus 4.6, GPT 5.4, and DeepSeek V4 Pro.
We run 100 iterations per experiment with reflective accumulation every five iterations.
At each iteration, \methodname can output one hypothesis.
A hypothesis is accepted only if the effect size $|\delta| \geq 0.2$ and $p \leq 0.05$ on both training and held-out validation splits, with an overfitting check requiring $|\delta_\text{val}| \geq 0.6 \cdot |\delta_\text{train}|$.
We allow \methodname{} to call other LLMs or vision models in evaluating proposed hypotheses, so that it can freely explore multimodal datasets.
Specifically, when calling vision language models (VLMs) to identify visual attributes from images it uses the same base LLM for the vision model. %
Meaning that if the default LLM is Claude Sonnet 4.6, then it will use the same model for reading images.
We run each experiment three times to calculate the mean and standard deviation.
Please refer to Appendix~\ref{sec:additional_impl_dets} for additional implementation details.

\noindent \textbf{Benchmark data.}    
Existing evaluation protocols for scientific discovery are inadequate for our open-ended systems.
Edge-recovery metrics (SHD, edge F1) cannot evaluate complex interactions such as  mediation chains or confound structures.
Question-answering benchmarks tell the agent what to look for, making them incompatible with the $\mathcal{P} = \emptyset$.
Therefore we construct two new multimodal benchmarks which target open-ended scientific discovery.  
The goal is to discover ecological patterns (\ie relationships, trends, associations, \etc) using data sourced from research-grade observations from the citizen science platform iNaturalist~\cite{iNaturalist}. 
Each observation on iNaturalist contains an image, latitude and longitude, positional accuracy, date of observation, and species name and taxonomic hierarchy. 
This makes it an ideal candidate for evaluating open-ended discovery from multimodal data.
We also source a set of ecological patterns from the academic literature which could be supported by the data, \eg ``Monarch Butterfly observations show northward latitude shifts during spring-summer, consistent with their documented migration corridors''~\citep{brower1996monarch}.
We create two datasets from this data: 
(i) \emph{iNatDisco-800} which has 800 observations across eight species and has nine ecological patterns obtained from peer-reviewed literature and 
(ii) \emph{iNatDisco-50K} which has 50,000 observations spanning 9,776 species with twelve ecological patterns. 
Additional dataset construction details can be found in Appendix~\ref{app:inatdisco}.

\begin{table}[t]
\centering
\caption{Open-ended discovery results on iNatDisco. \emph{Recall}: fraction of peer-reviewed ground truth patterns rediscovered. \emph{Support rate}: fraction of all proposed hypotheses that pass held-out statistical validation, not applicable to causal methods (--) as they output a fixed graph rather than iteratively proposing hypotheses. Higher scores are better for both metrics.}
\label{tab:inat}
\vspace{-.5em}
\footnotesize
\resizebox{1.0\textwidth}{!}{
\begin{tabular}{l l c ccccc}
\toprule
\textbf{Dataset} & \textbf{Method} & \textbf{Type} & $\mathcal{H}$ & \textbf{Reflect} & \textbf{Prior Info} & \textbf{Recall} & \textbf{Support Rate} \\
\midrule

\multirow{7}{*}{iNatDisco-800}
& LLM+PC        & \multirow{3}{*}{\rotatebox{90}{Causal}} & \Hedge     & \cross      & $\emptyset$ & 0/9  & -\\
& LLM+NOTEARS   &                                                  & \Hedge     & \cross      & $\emptyset$ & 0/9  & -\\
& GPT-4 BFS     &                                                  & \Hedge     & \cross      & $\emptyset$ & 1/9  & -\\
\cline{2-8}
& HeurekaBench~\cite{panigrahi2026heurekabench}& \multirow{4}{*}{\rotatebox{90}{LLM}} & \Hguidecode & \cross& Partial & 3/9 & 62.2\%$\pm$6\%\\
& ExperiGen~\citep{gupta2026accelerating}     &                                                  & \Hguidecode & \greencheck & Partial    & 3/9  & 56.6\%$\pm$5\%\\
& \methodname{} w/o Reflect &                                     & \Hopencode & \cross      & $\emptyset$ & 7/9 & 70.0\%$\pm$2\% \\
& \methodname{} (Ours) &                                          & \Hopencode & \greencheck & $\emptyset$ & \textbf{8/9} & \textbf{72.7\%$\pm$3\%} \\

\midrule

\multirow{7}{*}{iNatDisco-50K}
& LLM+PC        & \multirow{3}{*}{\rotatebox{90}{Causal}} & \Hedge     & \cross      & $\emptyset$ & 0/12 & - \\
& LLM+NOTEARS   &                                                  & \Hedge     & \cross      & $\emptyset$ & 1/12 & -\\
& GPT-4 BFS     &                                                  & \Hedge     & \cross      & $\emptyset$ & 1/12 & -\\
\cline{2-8}
& HeurekaBench~\cite{panigrahi2026heurekabench}& \multirow{4}{*}{\rotatebox{90}{LLM}} & \Hguidecode & \cross & Partial & 2/12 & 64.7\%$\pm$4\%\\
& ExperiGen~\citep{gupta2026accelerating}     &                                                  & \Hguidecode & \greencheck & Partial  & 3/12 & 67.8\%$\pm$5\%\\
& \methodname{} w/o Reflect &                                     & \Hopencode & \cross       & $\emptyset$ & 6/12 & 66.6\%$\pm$3\% \\
& \methodname{} (Ours)      &                                     & \Hopencode & \greencheck & $\emptyset$ & \textbf{8/12} & \textbf{74.2\%$\pm$3\%}  \\

\bottomrule
\end{tabular}
}
\vspace{-1.5em}
\end{table}

\subsection{Evaluation of open-ended discovery on iNatDisco}
\label{sec:exp_inat}
Here we demonstrate that \methodname{}, by combining code-driven hypothesis testing with meta-reflection, achieves higher performance on real world scientific discovery. 
We evaluate using the following metrics: (i) the recall of patterns discovered from the peer-reviewed literature annotations and (ii) the support rate of the proposed hypothesis.
The support rate, defined as the fraction of proposed hypotheses that pass held-out statistical validation, shows how well \methodname is able to open-endedly propose new ideas that can be validated on real datasets via code.
However, this metric can be hacked as \methodname  could apply \emph{p-hacking} by proposing `easy/obvious' hypotheses to obtain p-values that  pass validation.
Therefore the recall rate of the patterns previously discovered in peer-reviewed literature serves as an additional metric that validates the usefulness of the discoveries

The results are shown in Table~\ref{tab:inat}.
Classical causal discovery methods (LLM+PC, LLM+NOTEARS, GPT-4 BFS), even when equipped with an LLM to extract variables from raw data, recover at most 1 of 9 patterns on iNatDisco-800: their edge-level hypothesis space cannot express the interaction effects, mediation chains, and multi-variable ecological patterns that dominate our ground truth.
Guided LLM methods (HeurekaBench-like, ExperiGen-like) perform better at 3/9, but their search is constrained by the partial prior information we supply, limiting exploration beyond the seeded directions.
\methodname{} discovers 8 of 9 patterns on iNatDisco-800 with a support rate of 72.7\%, meaning nearly three-quarters of all hypotheses the system proposes are validated on held-out data.
On iNatDisco-50K, which contains 12 patterns across 9,776 species, \methodname{} recovers 8/12 patterns with an even higher support rate of 74.2\%, suggesting that larger datasets enable the system to ground its hypotheses more reliably even as the search space grows.
Compared to the ablation without \textsc{Reflect}, we observe consistent drops in both recall (7/9$\to$8/9 on iNatDisco-800, 6/12$\to$8/12 on iNatDisco-50K) and support rate, confirming that \textsc{Reflect} not only broadens what the system investigates but also improves the quality of its proposals.

\begin{figure}[t]
    \centering
    \includegraphics[width=0.9\linewidth]{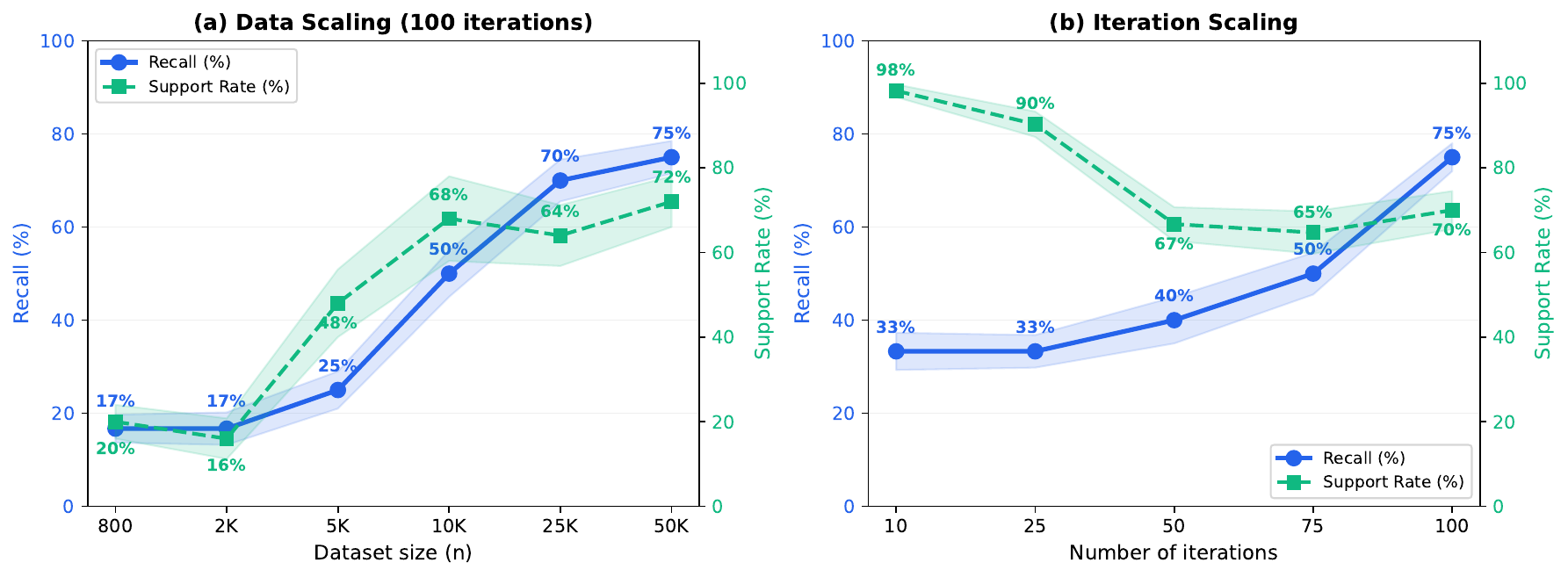}
    \vspace{-1.0em}
    \caption{Scaling behavior on iNatDisco-50K.
\textbf{(a)} Providing more data improves recall and yields more supported insights. 
\textbf{(b)} More model iterations increases recall but the support rate decreases as the model moves on from easy hypotheses and starts to propose more speculative ones.}
    \label{fig:scaling}
    \vspace{-1.5em}
\end{figure}

In Fig.~\ref{fig:scaling}, we further investigate the scaling behavior of \methodname{} w.r.t dataset size and the number of iterations.
We take iNatDisco-50K data and subsample it to create data subsets of different scales to probe the scaling behavior of \methodname w.r.t dataset size.
We see that more data makes subtle cross-kingdom patterns more discoverable, and \methodname{} is able to recover more supported patterns from the data (Fig.~\ref{fig:scaling} (a)).
Additionally, we investigated the behavior when scaling the number of iterations, where each iteration proposes and evaluates one hypothesis and \textsc{Reflect} runs every 5 iterations to analyze the claim store and steer subsequent proposals.
Increasing the number of iterations reduces the support rate because the model proposes increasingly speculative hypotheses after early iterations have exhausted the easy ones (Fig.~\ref{fig:scaling} (b)).
In contrast, recall consistently increases as we increase the number of iterations, suggesting that the \textsc{Reflect}-guided exploration continues to surface novel patterns even as per-hypothesis success rates decline.

\subsection{Counterfactual evaluation}
\label{sec:counterfact_results}

The ground truth patterns in iNatDisco are drawn from the peer-reviewed ecology literature, and thus we cannot guarantee they were absent from the base LLMs' training data. 
A system that simply recalls memorized facts, rather than testing hypotheses against data, would score well on this benchmark despite lacking genuine discovery capabilities.
To address this, we constructed iNatDisco-800-CF, a counterfactual variant in which five well-known ecological relationships are deliberately reversed in the tabular data while the image set remains unchanged.
Concretely, we modify observation timestamps, coordinates, and species-level metadata in the data table so that the original patterns no longer hold and the opposite relationships emerge instead.
For example, we fix bird latitude to a constant value regardless of month (removing seasonal migration), resample 70\% of fungal observations into spring months (reversing the autumn fruiting pattern), and narrow mammal geographic ranges while widening insect ranges (inverting the body-size and range relationship).
As the images are not altered, the VLM still sees the same species in the same habitats, but the tabular data now contradicts real-world ecology.
If a system relied only on LLM priors, it would report the real-world patterns (\eg ``fungi peak in autumn''). 
Instead, if it relies on data, it would report the counterfactual patterns actually present in the modified data.
See Appendix~\ref{app:counterfactual} for a detailed description of the specific data modifications we applied.

\begin{figure}
    \centering
    \includegraphics[width=0.9\linewidth]{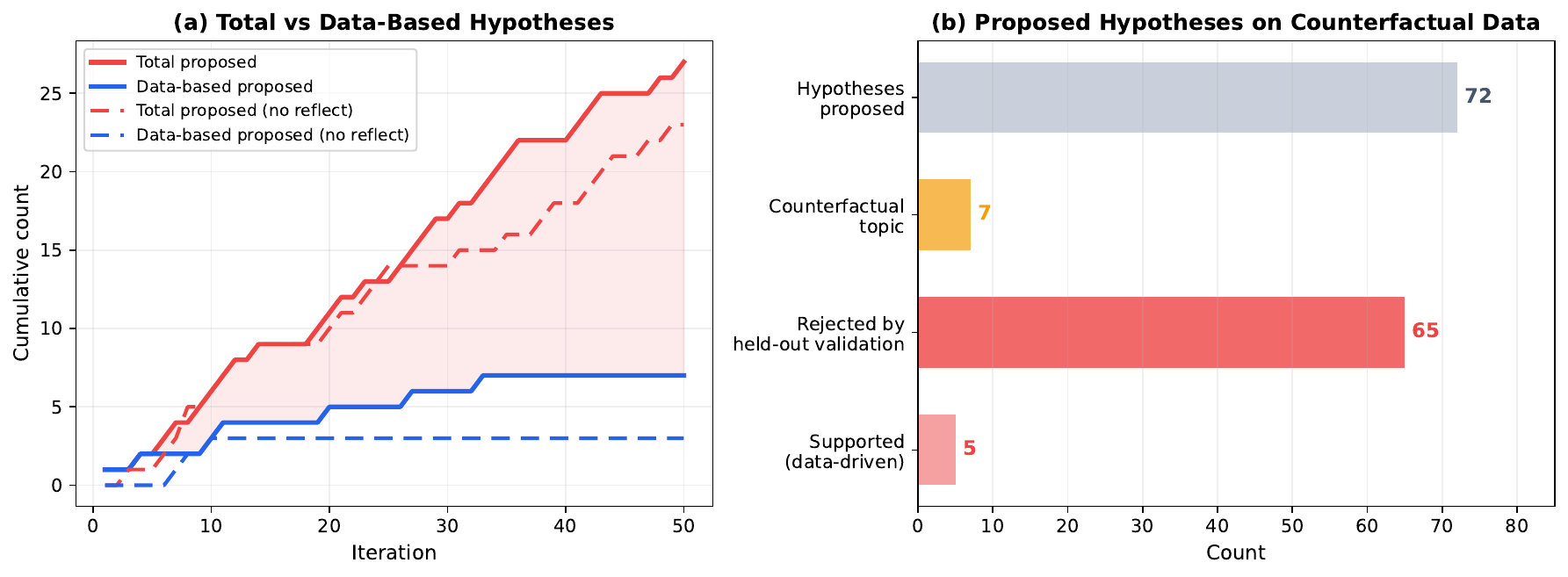}
    \vspace{-1.0em}
    \caption{Experiments on our iNatDisco-800-CF  counterfactual dataset. 
    \textbf{(a)} Cumulative number of proposed hypotheses over 50 iterations, separated into all hypotheses and data-based hypotheses, with and without \textsc{Reflect}. \textsc{Reflect} increases the number of hypotheses grounded in the observed input data rather than only in the model's priors.
\textbf{(b)} Distribution of proposed hypotheses. Most proposed hypotheses are rejected by held-out validation, the supported discoveries are data-driven and follow the counterfactual patterns present in the modified data.}
    \label{fig:counterfactual}
    \vspace{-1.2em}
\end{figure}

In Fig.~\ref{fig:counterfactual} (a) we classify the proposed hypotheses into whether or not they are based on the observed data  by examining if the \textsc{Propose} process involves tool calling that examines the dataset itself. 
The gap between the data-based hypotheses and the total number of hypotheses can be roughly seen as the number of hypotheses that the \textsc{Propose} module proposed mainly based on the LLM's internal knowledge.
We observe that as the number of iterations increases, the LLM keeps proposing hypotheses that originate from its internal knowledge and a smaller fraction of the proposed hypotheses are based on the data.
Additionally, the \textsc{Reflect} module results in more data-driven hypotheses than without, demonstrating its efficacy at helping to learn from past hypotheses and their tested validity.
In Fig.~\ref{fig:counterfactual} (b) we show the distribution of  hypotheses proposed by \methodname{}. 
Although \methodname{} proposes many hypotheses that do not have a basis in the data, the \textsc{Evaluate} process rejects ones that are not supported by data. 
Therefore the final supported discoveries are grounded in data, even if the data is counterfactual w.r.t. the LLM's internal knowledge.

\begin{figure}
    \centering
    \includegraphics[width=.9\linewidth]{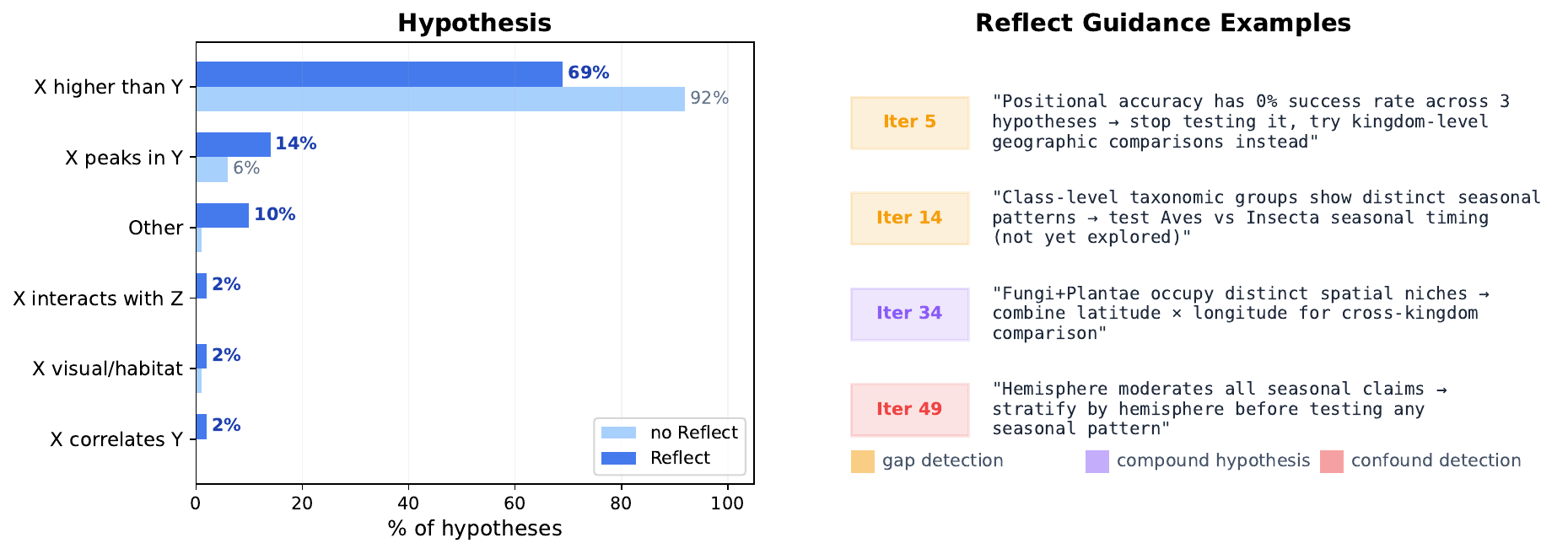}
    \vspace{-1.0em}
    \caption{Ablation of reflection. 
\textbf{Left:} distribution of generated hypothesis with and without \textsc{Reflect}. Without reflection, hypotheses are dominated by simple pairwise comparisons, while \textsc{Reflect} produces a broader set of seasonal, interaction, visual, and correlation-based hypotheses.
\textbf{Right:} examples of guidance produced by \textsc{Reflect}, including gap detection, compound hypothesis generation, and confound detection. These guidance messages redirect later proposal steps toward under-explored variables and higher-order relationships.}
    \label{fig:behavior}
    \vspace{-1.5em}
\end{figure}

\vspace{-1em}
\subsection{Ablations}
\label{sec:exp_ablation}
\vspace{-1em}

\begin{figure}[t]
    \centering
    \includegraphics[width=.9\linewidth]{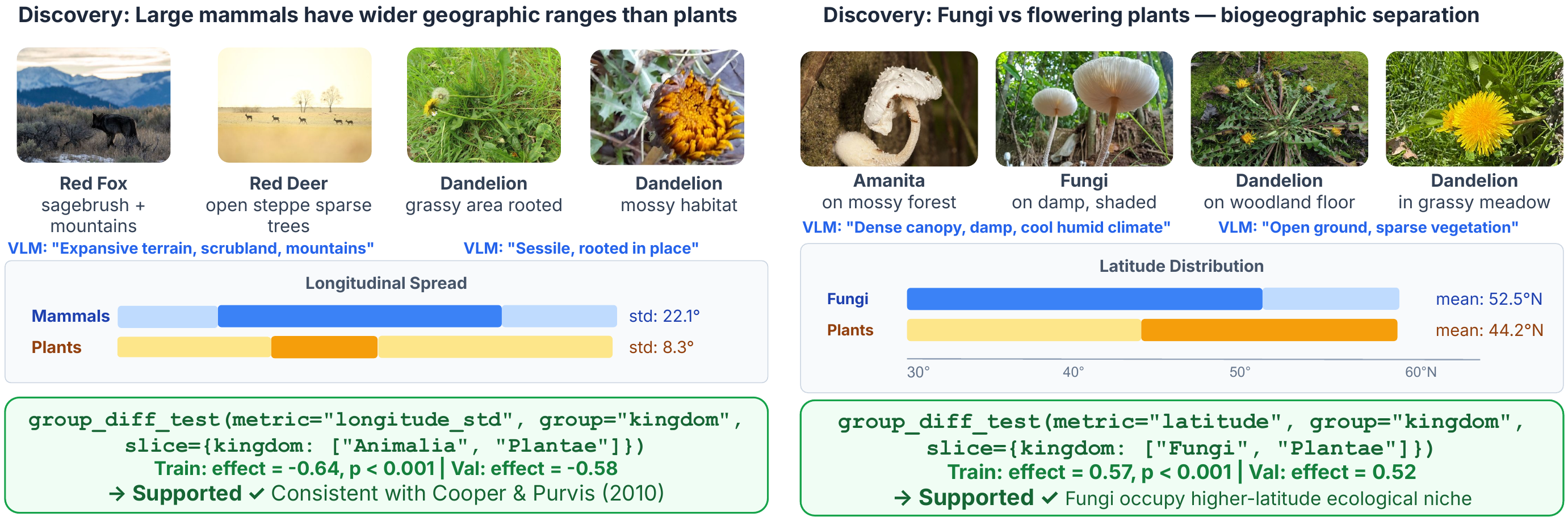}
    \vspace{-.5em}
    \caption{Examples of vision-grounded discoveries produced by \methodname{}.
\methodname{} can use visual evidence extracted from images to formulate and validate hypotheses that are not directly available from tabular metadata. 
\textbf{Left:} VLM-derived habitat descriptions support a discovery that mammals occupy broader longitudinal ranges than plants. 
\textbf{Right:} visual cues about canopy density and ground cover lead to a supported biogeographic separation between fungi and flowering plants, with fungi concentrated at higher latitudes. 
In both cases, the visual observations are converted into executable statistical tests and accepted only after validation on held-out data.}
    \label{fig:vlm}
    \vspace{-1.5em}
\end{figure}

We analyze the internal behavior of \methodname{} to explore how our \textsc{Reflect} module steers discovery and how vision  contributes to hypothesis generation. 
We also perform ablations on the backbone LLM and investigate the effect of optional user-provided context on discovery behavior.

\noindent \textbf{Reflection guidance.}
Fig.~\ref{fig:behavior} (right) illustrates concrete examples of \textsc{Reflect} guidance produced during a 100-iteration run on iNatDisco-50K.
The guidance falls into three categories, each addressing a distinct failure mode of unguided search:
\emph{Gap detection} identifies variables or relationships with zero coverage in the claim store.
At iteration~5, \textsc{Reflect} observes that positional accuracy has a 0\% support rate across attempted hypotheses and recommends abandoning it in favor of kingdom-level geographic comparisons, redirecting search away from an unproductive direction.
\emph{Compound hypothesis generation} combines insights from multiple claims to propose higher-order hypotheses.
At iteration~34, \textsc{Reflect} observes that Fungi and Plantae have been tested separately against latitude and longitude, and proposes testing their joint latitude$\times$longitude spatial niche separation.
\emph{Confound detection} flags variables that moderate multiple existing claims.
At iteration~49, \textsc{Reflect} notices that hemisphere appears as a moderator in all seasonal claims and recommends stratifying by hemisphere before testing any seasonal pattern, preventing it from reporting confounded associations.

Without \textsc{Reflect}, the distribution of hypotheses \methodname{} generates is heavily skewed toward simple pairwise comparisons (Fig.~\ref{fig:behavior} (left)): 92\% of base hypotheses follow the ``X is higher/greater than Y'' form, with no interaction or correlation hypotheses generated at all.
\textsc{Reflect} substantially diversifies the hypothesis space 
and reduces the dominance of direct pairwise comparison hypotheses from 92\% to 69\%, shifting \methodname{} toward more structured relational hypotheses. 
These include interaction hypotheses (0\% $\to$ 2\%), correlation tests (0\% $\to$ 2\%), and more specific seasonal or visual patterns, such as ``X peaks in Y'' (6\% $\to$ 14\%).
These shifts are modest in magnitude but important in effect, \eg the compound hypothesis at iteration~34 (``test Fungi$\times$Plantae spatial niche separation'') is one of only two interaction-type hypotheses in the entire run, yet it produces a statistically supported discovery ($p < 0.001$) that the baseline never formulates.

\noindent \textbf{Vision capabilities.}
\methodname{} is designed to support discovery of patterns in scientific databases beyond text or tabular metadata. By allowing our method to query a vision language model (VLM) using code, we access visual processing capabilities that enable \methodname{} to find patterns that require understanding information uniquely stored in the \textit{images} in our database. 
Fig.~\ref{fig:vlm} illustrates two example results.
In Fig.~\ref{fig:vlm} (left), the VLM describes mammal observations as occurring in expansive terrain such as scrubland and mountains, while plant observations are described as ``sessile and rooted in place''. 
These visual cues motivate \methodname to formulate a species range-size hypothesis comparing mammals and plants.
The resulting statistical test uses longitude information and finds that mammals have wider longitudinal ranges than plants in the data.
While the longitudinal spread itself can be computed from metadata alone, the visual descriptions provide ecological context for why this comparison is meaningful, \ie the system connects the geographic pattern to image-level cues about mobility, habitat openness, and rootedness.
Additionally in Fig.~\ref{fig:vlm} (right), the VLM observes fungi on ``mossy forest floors with dense canopy'' versus dandelions in ``open ground with sparse vegetation,'' motivating \methodname to formulate a biogeographic niche hypothesis comparing fungi and flowering plants. 
The resulting statistical test uses latitude metadata and finds that fungi occupy a higher-latitude niche than flowering plants.
Neither habitat type nor vegetation density appears in the metadata as both are extracted entirely from the images.

\vspace{-0.5em}
\section{Conclusion}
\vspace{-0.5em}

Autonomous scientific discovery requires more than proposing plausible hypotheses, \ie it requires open-ended exploration, empirical validation, and the ability to reason over accumulated evidence. 
We introduced \methodname, a code-driven discovery framework that formulates testable open-ended hypotheses from multimodal data, validates them on held-out splits, and periodically reflects over its own claims to identify gaps, confounds, and compounds derived insights. 
Across ecological and causal discovery benchmarks, this combination improves both the recall and the diversity of supported discoveries over systems constrained by fixed edge spaces, predefined questions, or short-term memory.
More broadly, our results suggest that progress in AI-assisted science depends on building agents that can organize an evolving body of evidence, not only generate isolated insights. 
\methodname is still limited by the scope and bias of the data it observes, and its discoveries require human scrutiny. 
Nevertheless, reflective, evidence-grounded agents offer a promising path toward scientific tools that help researchers surface overlooked patterns and formulate new testable questions.

\noindent{\bf Acknowledgements.} This work was in part supported by a Royal Society Research Grant, a Schmidt Sciences AI2050 Early Career Fellowship, an NSF CAREER Grant (Award No.~2441060), and the NSF and NSERC AI  Biodiversity Change Global Center (NSF Award No.~2330423 and NSERC Award No.~585136). 

\bibliographystyle{plain}
\bibliography{main}

\newpage
\appendix
\setcounter{table}{0}   
\renewcommand{\thetable}{A\arabic{table}}
\setcounter{figure}{0}
\renewcommand{\thefigure}{A\arabic{figure}}

{\LARGE Appendix}

\section{Additional results}
\label{sec:additional_results}

\subsection{Additional ablations}

\noindent \textbf{Base LLM comparison.}
Table~\ref{tab:ablations} (left) shows that \methodname is compatible with different backbone LLMs, but the choice of model substantially impacts discovery performance.
Claude Sonnet 4.5 achieves the highest recall, recovering 8/9  patterns with a 72.7\% support rate.
Claude Opus 4.6 obtains a slightly higher support rate (76.5\%) but recovers only 4/9 patterns, suggesting a more conservative search behavior that validates a larger fraction of its proposed hypotheses, but explores fewer of the benchmark patterns. 
GPT-5.4 and DeepSeek V4 Pro (without vision) recover 3/9 and 2/9 patterns, respectively.
The nonzero recall of the text-only DeepSeek variant indicates that the tabular analysis pipeline alone contains substantial signal, while the stronger vision-enabled models benefit from the additional image-derived evidence.

\noindent \textbf{Controllability.}
Table~\ref{tab:ablations} (right) shows that \methodname can be steered by providing  optional user context without modifying the underlying discovery loop.
In the prior-knowledge setting, the user provides only a small set of natural-language facts that should be treated as already known, such as ``fungi peak in autumn'' and ``monarchs migrate,'' without revealing the benchmark targets or expected discoveries.
In the guided setting, the user provides a single broad research interest, such as ``investigate observer bias in GPS accuracy,'' rather than a concrete hypothesis or test to execute.
Providing prior knowledge related to already-known facts increases topic adherence from 43\% to 54\% while preserving the same recall (7/9), indicating that the agent can incorporate external context without losing coverage of the benchmark patterns.
Providing a specific research focus produces a stronger steering effect, increasing topic adherence to 68\%, but reduces recall from 7/9 to 6/9 and lowers the support rate.
This reflects the expected trade-off of controllable open-ended search, \ie user context can redirect the system toward a desired region of the hypothesis space, but stronger steering narrows exploration and may reduce the number of broadly supported discoveries.

\begin{table}[h]
\centering
\caption{Model ablation on iNatDisco-800. \textbf{Left:} Recall across four LLMs.  $^\dagger$~denotes no vision used (\ie text only). \textbf{Right:} Impact of providing prior knowledge or a research focus to the system. Topic adherence measures the fraction of hypotheses related to the user-specified interest.}
\label{tab:ablations}
\footnotesize
\begin{minipage}[t]{0.40\textwidth}
\centering
\begin{tabular}{@{}lcc@{}}
\toprule
\textbf{Model} & \textbf{Supp.\ Rate} & \textbf{Recall} \\
\midrule
Sonnet 4.5              & 72.7\%            & 8/9 \\
Opus 4.6                & 76.5\%            & 4/9 \\
DeepSeek V4 Pro$^\dagger$   & 65.2\%            & 2/9 \\
GPT-5.4                 & 70.1\%            & 3/9 \\
\bottomrule
\end{tabular}
\end{minipage}
\hfill
\begin{minipage}[t]{0.56\textwidth}
\centering
\begin{tabular}{@{}lccc@{}}
\toprule
 & \textbf{Default} & \textbf{Prior Know.} & \textbf{Guided} \\
\midrule
User provides    & \emph{nothing} & \emph{known facts} & \emph{focus area} \\
Supp.\ Rate      & 72.7\% & 45.5\% & 36.7\% \\
Recall           & 8/9    & 7/9 & 6/9 \\
Topic adherence  & N/A   & 54\% & 68\% \\
\bottomrule
\end{tabular}
\end{minipage}
\end{table}

\subsection{Classical causal discovery benchmarks}
\label{app:classical}

Here we evaluate \methodname{} on two standard causal discovery benchmarks with validated causal DAGs: SACHS~\citep{sachs2005causal} (11 proteins, 17 edges) and ASIA~\citep{lauritzen1988asia} (8 variables, 8 edges). For each benchmark, we compare against classical structure learning methods (PC, GES, NOTEARS, DAG-GNN, GOLEM) on edge recovery, and report our method with and without \textsc{Reflect}.
Table~\ref{tab:edge_recovery} reports edge-recovery F1 against the published ground truth DAGs. 
\methodname{} achieves the highest F1 on SACHS (0.83) and is competitive on ASIA (0.86), outperforming all classical structure learning methods by a substantial margin.
On SACHS, the gap is particularly large: classical methods plateau at 0.33--0.48 F1, constrained by their edge-level hypothesis space, while \methodname{} can express and test richer relational patterns that map onto the same edges.
Removing \textsc{Reflect} consistently degrades performance (0.83$\to$0.72 on SACHS, 0.86$\to$0.80 on ASIA), confirming that reflective accumulation helps even on classical benchmarks by identifying under-tested variable pairs and reducing redundant proposals.

\begin{table}[h]
\centering
\caption{Edge recovery F1 on classical causal discovery benchmarks. The results for classical methods are obtained from the respective publications. Higher scores are better.}
\label{tab:edge_recovery}
\footnotesize
\begin{tabular}{@{}lcc@{}}
\toprule
\textbf{Method} & \textbf{SACHS} & \textbf{ASIA}  \\
\midrule
\methodname{}                                     & 0.83$\pm$0.03  & 0.86$\pm$0.05  \\
\methodname{} (no Reflect)                        & 0.72$\pm$0.04  & 0.80$\pm$0.03  \\
\midrule
PC~\citep{spirtes2000causation_PC}                & 0.48 & 0.77 \\
GES~\citep{chickering2002optimal_GES}             & 0.42 & 0.73 \\
NOTEARS~\citep{zheng2018NOTEARS}                  & 0.36 & 0.68 \\
DAG-GNN~\citep{yu2019dag}                         & 0.33 & 0.65 \\
GOLEM~\citep{ng2020role}                          & 0.38 & 0.69 \\
GPT-4 BFS~\citep{jiralerspong2024efficient_bfs}   & 0.74 & 0.93 \\
\bottomrule
\end{tabular}
\end{table}

\subsection{Synthetic visual benchmark}
\label{app:synthetic}

The iNatDisCo benchmarks evaluate discovery on real-world data, but they cannot isolate the contribution of vision from metadata as every visually-grounded hypothesis could potentially be proposed by an LLM that has memorized ecological knowledge. 
To disentangle these factors, we construct a synthetic benchmark where (i)~the visual features are novel (colored shapes on backgrounds, not real species), ensuring the LLM has no prior knowledge about the patterns, and (ii)~some ground truth patterns are only discoverable by analyzing the images (\eg ``red shapes are more common in autumn''), providing a controlled test of visual feature extraction.

\begin{figure}[h]
\centering
\includegraphics[trim={0 0 0 15pt},clip,width=0.9\linewidth]{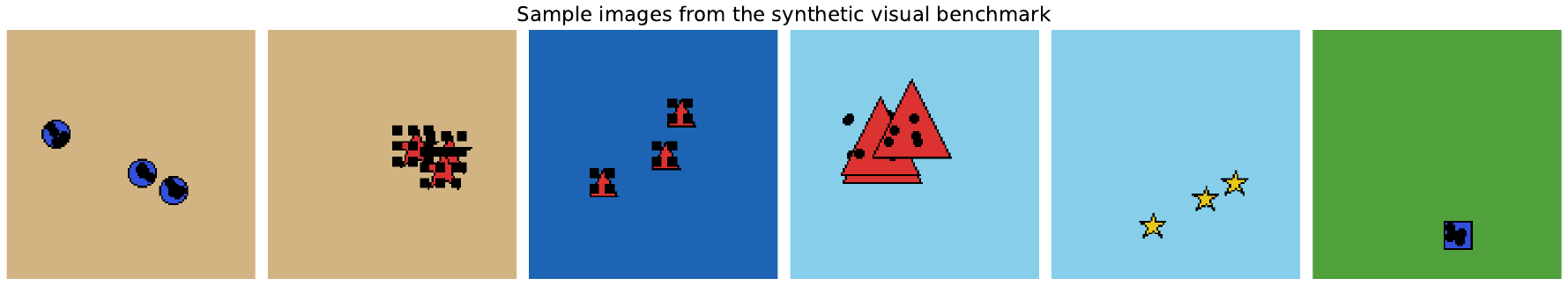}
\caption{Sample images from the synthetic visual benchmark. Shapes vary in color, size, and texture; backgrounds vary between sky, grass, water, and sand.}
\label{fig:synthetic_examples}
\end{figure}

We construct a synthetic benchmark to test \methodname{}'s ability to extract visual features from images for hypothesis testing.
The dataset contains 5{,}000 programmatically generated images of colored shapes on backgrounds. 
Fig.~\ref{fig:synthetic_examples} illustrates some representative examples. 
Each image has visual variables (\ie color, shape, size, texture, background, count) and tabular metadata (\ie category, region, season, temperature, elevation). 
Eight ground truth patterns span metadata-only, vision-only, and cross-modal relationships are constructed (see Table~\ref{tab:synthetic_gt}).

Results for \methodname{} on the synthetic dataset are presented in Table~\ref{tab:synthetic_results}.
At 100 iterations, the system recovers 3 of 8 ground truth patterns with a support rate of 54.2\%.
Both metadata-only patterns (P1, P2) are reliably discovered, confirming that the code-driven pipeline handles standard tabular relationships well.
One vision-only pattern is recovered, where the agent identifies the color-season association (P3) by leveraging the VLM during hypothesis generation.
However, the remaining vision-only and cross-modal patterns are not recovered: the agent often proposes correct visual hypotheses (\eg ``large shapes appear more often on grass backgrounds'') but the current vision tool pipeline lacks the statistical power to validate them on held-out data, as individual image classification followed by a chi-squared test introduces substantial noise.
This result highlights that the bottleneck for multimodal discovery is not hypothesis generation but statistical validation of visual features, motivating future work on richer vision-to-tabular feature extraction.

\begin{table}[h]
\centering
\caption{Ground truth patterns in the synthetic visual benchmark.}
\label{tab:synthetic_gt}
\footnotesize
\begin{tabular}{@{}clll@{}}
\toprule
\textbf{\#} & \textbf{Pattern} & \textbf{Type} & \textbf{Description} \\
\midrule
P1 & Temperature $\sim$ elevation & Metadata & $r = -0.7$ \\
P2 & Category A warmer & Metadata & Higher mean temperature \\
P3 & Red in autumn & Vision & 67\% of autumn images are red (vs 25\%) \\
P4 & Size $\sim$ background & Vision & Large on grass (61\%), small on water \\
P5 & Shape-texture exclusion & Vision & Circles/squares never striped \\
P6 & Blue $\sim$ elevation & Cross-modal & Blue at elevation 2402 vs 1484 \\
P7 & Region $\sim$ size & Cross-modal & North: 72\% large; South: 18\% \\
P8 & Spotted + cat.\ C $\to$ cold & Cross-modal & Anomalously low temperature \\
\bottomrule
\end{tabular}
\end{table}

\begin{table}[h]
\centering
\caption{Synthetic visual benchmark results.}
\label{tab:synthetic_results}
\footnotesize
\begin{tabular}{@{}lcccc@{}}
\toprule
\textbf{\# Iterations} & \textbf{Method} & \textbf{Type of Pattern} & \textbf{Supp.\ Rate} & \textbf{Recall}   \\
\midrule
- & HeurekaBench & All & 34.5\% & 0/8 \\
- & ExperiGen & All & 41.1\% & 1/8 \\
\midrule
50 iter    & Ours & All & 56.4\% & 2/8  \\
100 iter   & Ours & All & 54.2\% & 3/8  \\
\midrule
100 iter   & Ours & Metadata    & --  & 2/8 \\
100 iter   & Ours & Vision      & --  & 1/8 \\
100 iter   & Ours & Cross-Modal & --  & 0/8 \\
\bottomrule
\end{tabular}
\end{table}

\section{Additional dataset details}

\subsection{iNatDisco benchmark construction}
\label{app:inatdisco}

We construct two ecological discovery benchmarks from iNaturalist research-grade observations, we use the same set of images as the INQUIRE dataset~\cite{vendrow2024inquire}: 
(i) iNatDisco-800, containing 800 observations across 8 species with 9 ground-truth patterns, and 
(ii) iNatDisco-50K, containing 50{,}000 observations across 9{,}776 species with 12 ground-truth patterns.
Ground-truth patterns were curated from peer-reviewed ecological literature by identifying relationships that should be detectable from citizen-science observation data. 
For each candidate pattern, we tested whether the corresponding signal was present in our collected iNaturalist observations using the statistical tests available to the discovery system. 
We retained only patterns that were statistically supported in the dataset ($p < 0.05$, effect size $\geq 0.2$), and annotated each retained pattern with its supporting reference of published paper. 

Table~\ref{tab:inat_gt_merged} describes the sourced ground-truth patterns for iNatDisco-800 and iNatDisco-50K.
To compute recall, we check whether \methodname proposes validated claims that recover these curated patterns using the same observation data and statistical tools available during benchmark construction.
Note that while some of the ground truth patterns are about specific species or have specific requirements, the data in iNaturalist cannot always have the level of specificity that satisfies it.
Therefore we use a slightly loose requirements for some ground truth patterns to validate them on iNaturalist.
The validated patterns are then used as the ground truth patterns for iNaturalist.

\begin{figure}[h]
\centering
\includegraphics[trim={0 0 0 15pt},clip,width=\linewidth]{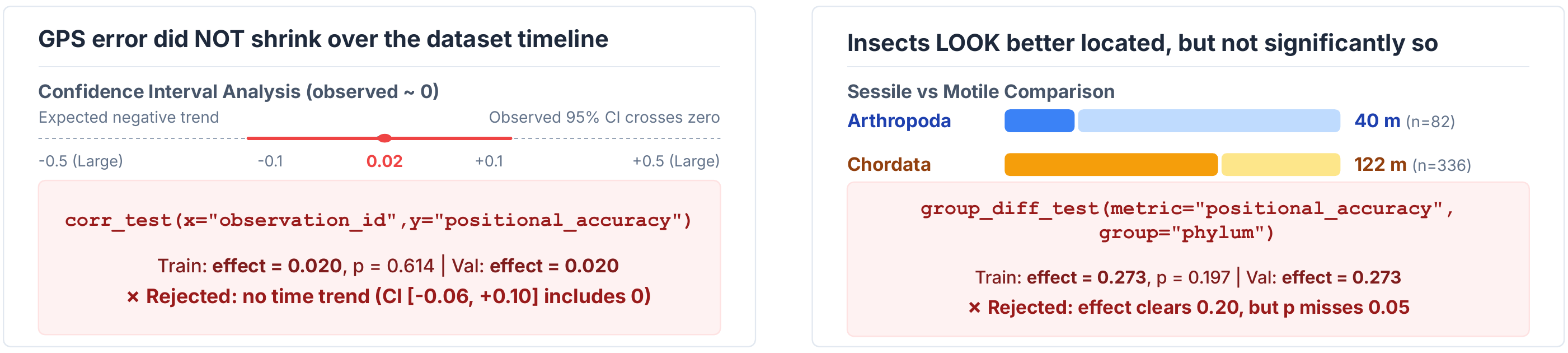}
\caption{Example rejected claims on iNatDisco.}
\label{fig:rejected_inatdisco}
\end{figure}

\begin{table}[h]
\centering
\caption{Ground-truth ecological patterns used in iNatDisco-800 and iNatDisco-50K. Three patterns are present in both datasets.}
\label{tab:inat_gt_merged}
\resizebox{\linewidth}{!}{
\begin{tabular}{@{}clp{3.2cm}p{4.8cm}p{4.8cm}@{}}
\toprule
\textbf{\#} & \textbf{Dataset} & \textbf{Pattern} & \textbf{Ecological Basis} & \textbf{Reference} \\
\midrule
1 & 800, 50K
& Fungi autumn fruiting
& \emph{A.\ muscaria} / fungal observations concentrated Sep--Nov
& ``Climate variation effects on fungal fruiting'' (Boddy et al., 2014)~\cite{boddy2014climate} \\

2 & 800
& Monarch migration
& \emph{D.\ plexippus} latitude increases Mar--Aug
& ``Monarch butterfly orientation: missing pieces of a magnificent puzzle'' (Brower, 1996)~\cite{brower1996monarch} \\

3 & 800
& Barn Swallow migration
& \emph{H.\ rustica} shifts northward in spring
& ``Climate change and the long-term northward shift in the African wintering range of the barn swallow'' (Ambrosini et al., 2011)~\cite{ambrosini2011climate} \\

4 & 800
& Red Fox wide range
& \emph{V.\ vulpes} spans 25--70\textdegree{}N
& ``Body size evolution in mammals: complexity in tempo and mode'' (Cooper and Purvis, 2010)~\cite{cooper2010body} \\

5 & 800
& Robin partial migration
& \emph{T.\ migratorius} northward shift spring--summer
& ``American Robin (\emph{Turdus migratorius}), version 1.0'' (Vanderhoff et al., 2020)~\cite{vanderhoff2020americanrobin} \\

6 & 800, 50K
& Plant seasonality / phenological offset
& Plantae concentrated in spring--summer; plants peak Apr--Jun vs animals May--Aug
& ``European phenological response to climate change matches the warming pattern'' (Menzel et al., 2006)~\cite{menzel2006european} \\

7 & 800
& Dandelion early flowering
& \emph{T.\ officinale} peaks Mar--May
& ``Seasonal variation in flowering of common dandelion'' (Gray et al., 1973)~\cite{gray1973seasonal} \\

8 & 800
& Red Deer northern habitat
& \emph{C.\ elaphus} concentrated 45--60\textdegree{}N
& ``Deer of the World: Their Evolution, Behaviour and Ecology'' (Geist, 1998)~\cite{geist1998deer} \\

9 & 800, 50K
& Hemisphere season inversion
& Seasonal patterns invert between NH and SH
& ``Response of tree phenology to climate change across Europe'' (Chmielewski \& R\"{o}tzer, 2001)~\cite{chmielewski2001response} \\

10 & 50K
& Latitudinal diversity gradient / insect tropical concentration
& Species richness increases toward the equator; insects more concentrated at $<$30\textdegree{} latitude
& ``On the generality of the latitudinal diversity gradient'' (Hillebrand, 2004)~\cite{hillebrand2004generality} \\

11 & 50K
& Bird latitudinal migration
& Birds shift northward spring--summer
& ``Adult male birds advance spring migratory phenology faster than females and juveniles across North America'' (Neate-Clegg et al., 2023)~\cite{neateclegg2023adult} \\

12 & 50K
& Amphibian spring emergence
& Amphibians peak sharply Mar--May
& ``The Ecology and Behavior of Amphibians'' (Wells, 2007)~\cite{wells2007ecology} \\

13 & 50K
& Lepidoptera wide latitude
& Butterflies span wider range than other insects
& ``The Structure and Dynamics of Geographic Ranges'' (Gaston, 2003)~\cite{gaston2003structure} \\

14 & 50K
& Fungi temperate concentration
& Fungi concentrated 40--60\textdegree{}N
& ``Global diversity and geography of soil fungi'' (Tedersoo et al., 2014)~\cite{tedersoo2014global} \\

15 & 50K
& Mammal temporal uniformity
& Mammals more uniform across months than birds
& ``Biogeography of time partitioning in mammals'' (Bennie et al., 2014)~\cite{bennie2014biogeography} \\

16 & 50K
& Continental endemism
& Certain families show continental endemism
& ``Biogeography,'' 4th ed.\ (Lomolino et al., 2010)~\cite{lomolino2010biogeography} \\

17 & 50K
& Elevation-latitude proxy
& Alpine plants at higher latitudes in mid-lat bands
& ``The use of `altitude' in ecological research'' (K\"{o}rner, 2007)~\cite{korner2007altitude} \\
\bottomrule
\end{tabular}
}
\end{table}

\subsection{iNatDisco-800-CF: counterfactual dataset construction}
\label{app:counterfactual}

Here we describe the iNatDisco-800-CF dataset that is used in Sec.~\ref{sec:counterfact_results} in the main paper. 
We construct iNatDisco-800-CF by modifying iNatDisco-800 to reverse five well-established ecological relationships (Table~\ref{tab:counterfactual}). For each relationship, we alter the underlying data so that the real-world pattern is no longer statistically present and the reversed pattern holds instead.

\begin{table}[b]
\centering
\caption{Counterfactual inversions in iNatDisco-800-CF. Each row shows the real-world relationship (which the LLM is expected to know), the counterfactual we inject into the data, and the modification applied.}
\label{tab:counterfactual}
\footnotesize
\begin{tabular}{@{}p{0.6cm}p{3.2cm}p{3.2cm}p{5cm}@{}}
\toprule
\textbf{\#} & \textbf{Real World} & \textbf{Counterfactual} & \textbf{Data Modification} \\
\midrule
CF1 & Birds migrate north in spring & Birds show no seasonal latitude shift & Set bird latitude to species mean $\pm$ $\mathcal{N}(0, 2)$ regardless of month \\
CF2 & Fungi fruit in autumn (Sep--Nov) & Fungi peak in spring (Mar--May) & Resample 70\% of fungal observations to months 3--5 \\
CF3 & Species richness increases toward equator & Richness decreases toward equator & Remove 60\% of observations at $|\text{lat}| < 30^\circ$ \\
CF4 & Large mammals have wide ranges & Mammals narrow, insects wide & Set mammal $\text{lat} \sim \mathcal{N}(\mu, 1.5)$; insect $\text{lat} \sim \mathcal{N}(\mu, 15)$ \\
CF5 & Plants peak in spring/summer & Plants peak in winter (Nov--Feb) & Resample 65\% of plant observations to months 11, 12, 1, 2 \\
\bottomrule
\end{tabular}
\end{table}

Each counterfactual was verified to be statistically present in the modified dataset. For example, after modification CF1, the Spearman correlation between bird latitude and month is $r = 0.02$, $p = 0.66$ (no significant seasonal shift). After CF2, 77\% of fungal observations fall in spring months. After CF4, mammal latitude standard deviation is 2.5$^\circ$ versus 14.8$^\circ$ for insects.

The counterfactual evaluation tests a critical property of code-driven discovery because every claim must pass held-out statistical validation, the system cannot ``hallucinate'' a pattern that is not in the data, even when the LLM's prior knowledge strongly suggests it should be there. As reported in Fig.~\ref{fig:counterfactual}, the LLM does propose prior-based hypotheses (\eg ``fungi peak in autumn'') but these are rejected by the held-out validation because the data shows the opposite.

\section{Additional implementation details}
\label{sec:additional_impl_dets}

\subsection{Evaluation protocol: LLM-as-a-judge for recall}
\label{app:llm_judge}

Evaluating open-ended discovery requires matching free-form natural-language claims against ground truth patterns that may be worded differently, use different species exemplars, or operate at a different level of generality.
String matching and keyword overlap fail in this setting, \eg a system claim ``Migratory insects shift northward in spring'' should match the ground truth ``Monarch butterfly latitude increases March--August'' even though they share few similar words.
We therefore use an LLM-as-a-judge to perform semantic matching.

\noindent \textbf{Input filtering.}
Only \emph{supported} claims are submitted to the judge.
Rejected and inconclusive claims are excluded.
This ensures that the recall metric measures not just whether the system \emph{proposed} a pattern, but whether it proposed a pattern \emph{and} produced code that validates it on held-out data.

\noindent \textbf{Batched scoring.}
We present the judge with the full list of ground truth patterns and one supported claim at a time.
For each claim, the judge outputs:
\begin{itemize}
    \item \texttt{best\_pattern\_id}: the ground truth pattern ID that best matches the claim, or \texttt{none} if no match exists.
    \item \texttt{score}: an integer on a 0--2 scale:
    \begin{itemize}
        \item 2 (exact match): the claim captures the same underlying ecological mechanism as the ground truth, even if worded differently or using different species exemplars (\eg ``Migratory insects shift northward in spring'' matches ``Monarch butterfly latitude increases March--August'').
        \item 1 (partial match): the claim is related but captures only part of the relationship, is too vague, or describes a consequence rather than the core pattern.
        \item 0 (miss): the claim describes a completely different phenomenon.
    \end{itemize}
    \item \texttt{reasoning}: a one-sentence explanation of the scoring decision.
\end{itemize}

The judge is instructed to focus on \emph{semantic equivalence of the underlying ecological mechanism}, not surface-level wording.
A generalized version of a ground truth pattern (\eg using a higher taxonomic level or a different exemplar species) receives score~2 if the core mechanism is identical.
Each claim may match at most one ground truth pattern.
The judge is also instructed not to match claims to prior-knowledge or obvious-tier patterns, ensuring that recall is computed only over novel patterns.

\noindent \textbf{Aggregation.}
For each ground truth pattern, we record the highest score received from any supported claim.
A pattern is considered ``discovered'' if its best score is $\geq 1$ (partial or exact match).
Recall is then the number of discovered pattern over the total number of peer-reviewed patterns.
We report recall as fractions (\eg 8/9) rather than percentages to make the denominator explicit.

\noindent \textbf{Judge model.}
We use Claude Sonnet 4.5 as the judge model via structured JSON output (function calling).
The same judge model is used across all experiments and baselines to ensure comparability.
We verified the judge reliability by manually inspecting all score-2 matches across our primary experiments (iNatDisco-800 and iNatDisco-50K).
In 95\% of cases, the judge's semantic matching agreed with manual assessment.

\subsection{LLM-based baseline construction}
\label{app:baselines}

We implement two LLM baselines as ablations of our framework by varying the prior information $\mathcal{P}$ provided to the system (see Table~\ref{tab:formalization} in the main paper).
A central claim of our work is that open-ended discovery (\ie $\mathcal{P} = \emptyset$) combined with reflective accumulation yields better results than concurrent guided approaches.
To test this rigorously, we implement ExperiGen-like and HeurekaBench-like baselines \emph{within our own framework}, ensuring that the only differences are (i)~the prior information $\mathcal{P}$ provided to the hypothesis generator and (ii)~whether \textsc{Reflect} is enabled.
All other components (\eg the backbone LLM (Claude Sonnet 4.5)), the statistical tool suite (Appendix~\ref{app:tools}), the held-out validation protocol (effect size $\geq 0.2$, $p \leq 0.05$ on both train and validation splits), the experiment planner, and the insight evaluator are identical across all conditions.
This controlled design isolates the effect of guidance and reflection from confounds due to model choice, tool availability, or evaluation criteria.

\noindent \textbf{HeurekaBench-like ($\mathcal{P} = \text{Full}$).}
HeurekaBench~\citep{panigrahi2026heurekabench} evaluates agents on pre-specified research questions with known ground-truth answers.
We simulate this by replacing the open-ended hypothesis generator prompt with nine specific research questions drawn from our iNatDisco-800 ground truth.
The agent is instructed to pick one unanswered question per iteration and design a rigorous statistical test for it.
\textsc{Reflect} is disabled, as HeurekaBench has no accumulation mechanism.
The full system prompt is:

\begin{tcolorbox}[colback=gray!5, colframe=gray!50, title=HeurekaBench-like Baseline Prompt]
\small
\texttt{You are analyzing a biodiversity dataset.}

\texttt{TASK DESCRIPTION: This dataset contains research-grade wildlife observations across species with GPS coordinates, dates, and taxonomic classification. The goal is to find statistically significant relationships between species traits, geographic distribution, and temporal patterns.}

\texttt{DATASET SUMMARY:}\\
\texttt{- 800 observations across 8 species}\\
\texttt{- Variables: latitude, longitude, month, season, kingdom, class\_name, species\_name, positional\_accuracy}\\
\texttt{- Species include: Vulpes vulpes, Cervus elaphus (mammals), Hirundo rustica, Turdus migratorius (birds), Danaus plexippus (insect), Amanita muscaria (fungi), Helianthus annuus, Taraxacum officinale (plants)}\\
\texttt{- Geographic range: global, with concentration in Northern Hemisphere}

\texttt{Generate hypotheses about ecological patterns in this data. Test each with appropriate statistical methods.}
\end{tcolorbox}

This setup provides the agent with \emph{maximum} guidance: it knows exactly what to look for.
Despite this advantage, the agent recovers only 3/9 patterns in the standard setting (see Table~\ref{tab:inat}), because (i)~many questions are phrased at the species level while the ground-truth patterns involve cross-taxon interactions, and (ii)~the fixed question list cannot adapt to emerging evidence.

\noindent \textbf{ExperiGen-like ($\mathcal{P} = \text{Partial}$).}
ExperiGen~\citep{gupta2026accelerating} provides agents with a task description, dataset summary, and seed hypotheses to accelerate discovery.
We simulate this by replacing the hypothesis generator prompt with a structured task description that includes the dataset name and size (800 observations, eight species across three kingdoms), all variable names and types, the species list with taxonomic classification, the geographic scope, and a general instruction to ``find statistically significant relationships between species traits, geographic distribution, and temporal patterns.''
\textsc{Reflect} is disabled, as ExperiGen uses implicit conditioning on prior hypothesis-evidence pairs rather than explicit meta-analysis.
The full system prompt is:

\begin{tcolorbox}[colback=gray!5, colframe=gray!50, title=ExperiGen-like Baseline Prompt]
\small
\texttt{You are analyzing a biodiversity dataset.}

\texttt{TASK DESCRIPTION: This dataset contains research-grade wildlife observations across species with GPS coordinates, dates, and taxonomic classification. The goal is to find statistically significant relationships between species traits, geographic distribution, and temporal patterns.}

\texttt{DATASET SUMMARY:}\\
\texttt{- 800 observations across 8 species}\\
\texttt{- Variables: latitude, longitude, month, season, kingdom, class\_name, species\_name, positional\_accuracy}\\
\texttt{- Species include: Vulpes vulpes, Cervus elaphus (mammals), Hirundo rustica, Turdus migratorius (birds), Danaus plexippus (insect), Amanita muscaria (fungi), Helianthus annuus, Taraxacum officinale (plants)}\\
\texttt{- Geographic range: global, with concentration in Northern Hemisphere}

\texttt{Generate hypotheses about ecological patterns in this data. Test each with appropriate statistical methods.}
\end{tcolorbox}

This setup provides \emph{partial} guidance: the agent knows the variable space and the general direction of inquiry, but must decide which specific hypotheses to pursue.
It recovers 3/9 patterns on iNatDisco-800 (see Table~\ref{tab:inat}), similar to the HeurekaBench-like baseline, because the generic task framing tends to produce simple pairwise comparisons rather than the compound cross-taxon patterns that dominate the ground truth.

\noindent \textbf{\methodname{} without reflection ($\mathcal{P} = \emptyset$, no Reflect).}
This baseline uses the same open-ended setting as our full system ($\mathcal{P} = \emptyset$) but with $\mathcal{G}_t = \emptyset$ for all $t$, \ie the hypothesis generator never receives guidance from the \textsc{Reflect} step.
This isolates the contribution of reflective accumulation while keeping all other components identical, including the open-ended system prompt that prioritizes cross-kingdom and interaction hypotheses.

\noindent \textbf{Controlling for fairness.}
We emphasize three design choices that ensure a fair comparison:
\emph{(i)} All baselines use the same backbone LLM, so differences in recall are not attributable to model capability.
\emph{(ii)} All baselines use the same held-out validation protocol, so differences in support rate reflect genuine differences in hypothesis quality, not evaluation stringency.
\emph{(iii)} The HeurekaBench-like baseline is given research questions that directly correspond to ground-truth patterns, an \emph{advantage} over our open-ended system, which must discover these patterns from scratch.
That our system achieves higher recall despite this disadvantage strengthens the case for open-ended exploration with reflective accumulation.

\subsection{Additional details}

\noindent \textbf{Classical causal discovery baselines.}
For the SACHS and ASIA benchmarks, we compare against classical structure learning algorithms: PC~\citep{spirtes2000causation_PC}, GES~\citep{chickering2002optimal_GES}, NOTEARS~\citep{zheng2018NOTEARS}, DAG-GNN~\citep{yu2019dag}, and GOLEM~\citep{ng2020role}. 
For PC and GES, we used the \texttt{causal-learn} Python package with default hyperparameters. 
For NOTEARS, DAG-GNN, and GOLEM, we report published results from their respective papers on the same SACHS and ASIA datasets.

Each classical method outputs a set of directed edges (\eg PKA $\to$ Raf). To evaluate these against our pattern-level ground truth, we convert each edge to a natural-language claim of the form ``\texttt{[Variable A]} has a direct causal effect on \texttt{[Variable B]}.'' These converted claims are then scored by the same LLM judge used for all other methods, ensuring a fair comparison on identical ground truth.

\noindent \textbf{Variable extraction for causal baselines.}
On SACHS, the 11 protein variables (Raf, Mek, Erk, Akt, PKA, PKC, P38, JNK, Plcg, PIP2, PIP3) are used directly as nodes in the causal graph. On ASIA, the 8 binary variables (asia, tub, smoke, lung, bronc, either, xray, dysp) are used directly. 
No feature engineering or variable selection is applied, the methods receive the raw variables as defined in the original benchmark publications.

\subsection{Statistical tools}
\label{app:tools}
The \textsc{Evaluate} step executes hypothesis code using a fixed set of seven statistical primitives. 
The hypothesis generator selects which tool to invoke and specifies its parameters; no manual intervention is required. Table~\ref{tab:tools} describes each tool.

\begin{table}[h]
\centering
\caption{Statistical tools available to \methodname{}. The first five operate on tabular data and the last two use a vision language model to extract visual features from images before applying statistical tests.}
\label{tab:tools}
\footnotesize
\begin{tabular}{@{}lp{9cm}@{}}
\toprule
\textbf{Tool} & \textbf{Description} \\
\midrule
\texttt{corr\_test} & Tests the correlation between two continuous variables using Spearman (rank-based, non-parametric) or Pearson (linear) correlation. Returns the correlation coefficient as effect size and the associated $p$-value. \\
\texttt{group\_diff\_test} & Tests whether a continuous metric differs between exactly two groups. Computes Cliff's delta (non-parametric) or Cohen's $d$ (parametric) as effect size, with significance assessed via permutation test ($n=1{,}000$ shuffles), Mann--Whitney $U$, or Student's $t$-test. \\
\texttt{predictive\_test} & Trains a random forest classifier to predict a target variable from a set of features, evaluated via 5-fold cross-validation. Returns AUC as effect size, testing whether the features carry predictive signal about the target. \\
\texttt{cluster\_and\_enrich} & Applies $k$-means clustering to the feature space, then tests whether cluster membership is associated with a categorical variable using a chi-squared test. Returns Cram\'{e}r's $V$ as effect size. \\
\texttt{stratified\_retest} & Re-runs a primary test (\eg \texttt{corr\_test}) separately within each stratum of a stratifying variable. Reports per-stratum results and a stability score measuring how consistently the effect holds across strata. Used to test for confounds and interaction effects. \\
\midrule
\texttt{visual\_attribute\_test} & Sends individual images to a VLM, which classifies a specified visual attribute (\eg habitat type: forest / grassland / urban). The resulting labels are tested against a metadata grouping variable using a chi-squared test with Cram\'{e}r's $V$ as effect size. \\
\texttt{visual\_group\_comparison} & Sends images from two groups to a VLM simultaneously and asks it to identify systematic visual differences. Returns a structured list of differences with per-difference confidence scores; the mean confidence serves as the effect size. \\
\bottomrule
\end{tabular}
\end{table}

All tools return a standardized result dictionary containing \texttt{effect\_size}, \texttt{p\_value}, and \texttt{status}. 
A hypothesis is accepted into the claim store only if the effect size exceeds $0.2$ and $p \leq 0.05$ on both the training and held-out validation splits.

\subsection{Prompts}
\label{app:prompts}

Here we provide the full prompts used in each component of \methodname{}.
The same prompts are used for iNatDisco datasets as well as the causal discovery and synthetic datasets.

\subsubsection{Hypothesis generation (\textsc{Propose})}
The hypothesis generator receives the dataset summary, prior knowledge, recent discovery results, and a list of hypotheses to avoid. 
The recent discovery results and a list of hypotheses to avoid are initialized to none at the start.
On subsequent iterations, the prompt includes the claim store and guidance from \textsc{Reflect}.

\begin{tcolorbox}[colback=gray!5, colframe=gray!50, title=Hypothesis Generation Prompt]
\small
You are a scientific hypothesis generator for ecological data analysis.

Your task is to generate a NOVEL testable hypothesis about patterns in the provided dataset.

\textbf{Dataset Summary:} \{dataset\_summary\}

\textbf{Prior Knowledge (DO NOT re-test):} \{prior\_knowledge\}

\textbf{Recent Discovery Results:} \{recent\_insights\}

\textbf{Hypotheses to AVOID:} \{hypotheses\_to\_avoid\}

Instructions:
\begin{enumerate}
\item Propose ONE specific, testable hypothesis that is DIFFERENT from previous attempts
\item If previous hypotheses about a topic were rejected, try a completely different angle
\item Focus on unexplored variable combinations
\item Look for interaction effects, threshold effects, or conditional relationships
\end{enumerate}

Output a JSON object with: statement, scope, variables, expected\_direction, risk\_flags.
\end{tcolorbox}

When images are available, the prompt is augmented with:

\begin{tcolorbox}[colback=blue!3, colframe=blue!30, title=Visual Hypothesis Augmentation]
\small
\{n\_images\} sample images from the dataset are attached. These are the ACTUAL ecology scene images. Look at them to understand:
\begin{itemize}
\item What visual elements vary across scenes (vegetation, water, animals, terrain, sky)
\item How those visual elements might relate to the numeric features
\item Patterns you can SEE that the numeric features might not fully capture
\end{itemize}
Your hypothesis should be informed by what you SEE in the images, not just the numbers.
\end{tcolorbox}

\subsubsection{Experiment planning (\textsc{Evaluate})}

The experiment planner receives the hypothesis and selects a statistical tool with appropriate parameters.

\begin{tcolorbox}[colback=gray!5, colframe=gray!50, title=Experiment Planning Prompt]
\small
You are an ecological statistician planning experiments.

CRITICAL RULES FOR CODE-CLAIM ALIGNMENT:
\begin{enumerate}
\item The grouping variable MUST match what the hypothesis compares (species $\to$ species\_name, class $\to$ class\_name, kingdom $\to$ kingdom)
\item The metric MUST measure what the claim describes (``seasonal shift'' = difference in latitude between seasons, not raw latitude)
\item The data slice MUST include exactly the populations the claim describes
\item If the hypothesis is about a visual property, use visual\_attribute\_test
\end{enumerate}

Available tools: corr\_test, group\_diff\_test, visual\_attribute\_test, visual\_group\_comparison, predictive\_test, stratified\_retest.

Output a JSON object with: method, feature\_spec, dataset\_slice\_spec.
\end{tcolorbox}

\subsubsection{Reflective accumulation (\textsc{Reflect})}

The \textsc{Reflect} agent receives all accumulated claims and produces structured guidance.

\begin{tcolorbox}[colback=orange!5, colframe=orange!40, title=Meta-Insight Discovery Prompt]
\small
You are a meta-science analyst. Your task is to discover META-INSIGHTS.

A Meta-Insight is NOT another finding about raw data. It is an OBSERVATION ABOUT THE PATTERN OF INSIGHTS that provides ACTIONABLE GUIDANCE for future hypothesis generation.

Types of meta-insights:
\begin{itemize}
\item \textbf{Confound Pattern:} A variable appears across multiple claims, suggesting it should be controlled
\item \textbf{Variable Cluster:} Variables that always co-occur in supported claims, suggesting interaction effects
\item \textbf{Gap:} Variables or relationships with zero coverage in the claim store
\item \textbf{Success Pattern:} Certain hypothesis types have higher support rates than others
\item \textbf{Contradiction:} Two claims report opposite effects, requiring stratified analysis
\item \textbf{Interaction Hint:} Overlapping variables across claims imply a compound hypothesis
\end{itemize}

\textbf{Dataset Summary:} \{dataset\_summary\}

\textbf{Accumulated Base Insights:} \{meta\_analysis\}

Generate \{n\_meta\_insights\} meta-insights, each with: observation, actionable\_recommendation, affected\_variables, meta\_type, priority, source\_insight\_ids.
\end{tcolorbox}

The meta-insights are then used to generate guided hypotheses:

\begin{tcolorbox}[colback=orange!5, colframe=orange!40, title=Guided Hypothesis Prompt]
\small
You are a hypothesis generator. Generate NEW TESTABLE HYPOTHESES that are GUIDED BY META-INSIGHTS.

Meta-insights tell you: what confounds to control for, what variable interactions to explore, what gaps to fill, what contradictions to resolve.

\textbf{Meta-Insights:} \{meta\_insights\}

\textbf{Dataset Summary:} \{dataset\_summary\}

Generate \{n\_hypotheses\} hypotheses that directly address the meta-insights. Each hypothesis should cite which meta-insight it addresses.
\end{tcolorbox}

\section{Compute resources}
\label{app:compute}

\begin{table}[h]
\centering
\caption{Compute budget for all experiments.}
\label{tab:compute}
\footnotesize
\begin{tabular}{@{}lr@{}}
\toprule
\textbf{Resource} & \textbf{Amount} \\
\midrule
Total completed runs & $\sim$86 \\
Total API calls & $\sim$8{,}250 \\
Estimated input tokens & $\sim$33M \\
Estimated output tokens & $\sim$8M \\
Estimated API cost (Sonnet pricing) & $\sim$\$220 \\
Wall-clock time (all experiments) & $\sim$72 hours \\
\bottomrule
\end{tabular}
\end{table}

LLM API calls take the majority of the computing resources of this work.
Each iteration of \methodname{} involves approximately three LLM API calls: one for \textsc{Propose} (hypothesis generation), one for experiment planning, and one for \textsc{Reflect} (every $K$ iterations). 
A 50-iteration run completes in 30--60 minutes depending on model latency and a 100-iteration runs take 1--2 hours.
Table~\ref{tab:compute} summarizes the total compute budget for all experiments reported in the paper.
The multi-model experiments (Section~\ref{sec:exp_ablation}) used Claude Sonnet 4.5, Claude Opus 4.6, GPT-5.4, and DeepSeek V4 Pro via their respective APIs. 
Classical causal discovery baselines (PC, GES) ran locally in under one minute each using the \texttt{causal-learn} package.

\section{Limitations} 
\methodname does not control the input data generation process and thus cannot make discoveries if they are not supported by the data available. 
For example, there are known biases in citizen science data which means that it may not accurately reflect underlying ecological phenomena~\cite{johnston2023outstanding}. 
As a result, human verification of any proposed discoveries is essential.  

Our iNatDisco datasets contain claims that have been validated in the academic literature. 
However, the set of claims is not complete, \ie there will be other valid patterns that a model could propose that are not annotated in the data. 
Importantly, the claims output by \methodname are selected based on their passing a statistical significant test which ensures that they are grounded in evidence in the data. 
In Appendix~\ref{sec:additional_results} we also report results on other datasets where the full set of valid claims is known.

\section{Broader impact}
\label{app:broader_impact}

\methodname{} could accelerate discovery on large observational datasets (\eg citizen science platforms) where human analysis capacity is limited, complementing domain expertise. 
The held-out validation and counterfactual evaluation provide safeguards against hallucinated findings. 
However, system's outputs should be treated as candidates for expert review, not established facts.

\end{document}